\documentclass[runningheads]{llncs}
\usepackage[T1]{fontenc}
\usepackage{amsmath}
\usepackage{hyperref}
\usepackage{graphicx}
\usepackage[dvipsnames]{xcolor}
\usepackage{pgf-pie}     
\usepackage{tikz}
\usepackage{pgfplots}
\pgfplotsset{compat=1.18}
\begin{document}
\title{Why Ranking Anomaly Detection Algorithms Isn't as Reliable as You May Think}
\titlerunning{Why Ranking AD Algorithms Isn't as Reliable as You May Think}
%
\author{Simon Klüttermann\inst{1} \and
 Jérôme Rutinowski\inst{2,3} \and
 Frederik Polachowski\inst{4}, \\
 Alice Kirchheim\inst{2,3,5}}

 \authorrunning{Klüttermann et al.}
%
 \institute{Carnegie Mellon University, Pittsburgh, USA \and
 TU Dortmund University, Dortmund, Germany\\
 \and
Lamarr Institute for Machine Learning and Artificial Intelligence, Germany \and
 iits Consulting GmbH,  Dortmund, Germany
 \and
 Fraunhofer Institute for Material Flow and Logistics IML, Dortmund, Germany
  \email{sklutter@andrew.cmu.edu}\\
}
\maketitle              
%
\begin{abstract}
Anomaly detection is a safety-critical machine learning problem with applications ranging from fraud detection to network intrusion prevention and industrial monitoring. Despite the large number of proposed anomaly detection algorithms, many novel methods claim state-of-the-art performance.
However, many authors do so under benchmark settings that are not aligned with one another. This lack of comparability raises concerns regarding the reproducibility and reliability of anomaly detection benchmarks. 

In this work, we study the impact of common benchmarking choices on the stability of algorithm rankings. Using seven representative anomaly detection algorithms and 690 datasets from the OddBench benchmark suite, we analyze how rankings change under varying dataset selections, evaluation metrics, hyperparameter configurations, and random seeds. To quantify this effect, we introduce a rank instability metric measuring the variability of algorithm rankings across benchmark settings.

Our results show that algorithm rankings in anomaly detection are highly unstable. In many cases, almost every competitive algorithm can appear as the best-performing method under some benchmark configuration. Among the studied factors, dataset selection and hyperparameter choice contribute most strongly to ranking uncertainty, while random seeds and evaluation metrics have a comparatively limited impact. We also observe that reliable benchmarking requires substantially larger and more diverse dataset collections than the ones commonly used in prior work.

Based on these findings, we recommend that future anomaly detection benchmark studies use at least $200$ datasets and explicitly evaluate hyperparameter sensitivity. More broadly, our analysis suggests that minor improvements in benchmark rankings should be interpreted cautiously, as the notion of state-of-the-art performance in anomaly detection is often highly dependent on experimental design choices.
Only once major improvements on a large amount of datasets can be observed and are invariant to design choices, it can be assumed that a tangible improvement has been made.

\keywords{Anomaly Detection  
\and Outlier Detection 
\and Benchmarking 
}
\end{abstract}
%
%

\section{Introduction}

Anomaly Detection (AD) is a crucial machine learning problem with many safety-critical applications, ranging from fraud detection~\cite{fraudapl,auto_appl_electionfraud} and failure prevention~\cite{pm2carvalho2019systematic,faultapl2} to network intrusion detection~\cite{icmlaChairIntrusionDetection,auto_appl_networkintrusion}. Matching this practical relevance, a large variety of anomaly detection algorithms have been proposed over the years, including classical statistical and distance-based approaches~\cite{ifor,knn}, modern deep learning methods~\cite{DEAN,dte}, and more recently foundation model-based approaches~\cite{outformer,fomox}. 

Paradoxically, despite the large methodological diversity, many newly proposed approaches report superior performance over previous methods~\cite{outformer,sean,dte}, which, of course, cannot be true for all of them at once. As a consequence, selecting the ``best'' anomaly detection algorithm becomes increasingly difficult. While several survey papers and benchmarking studies exist \cite{macrodata,surveyzhao,myhyperparam,surveyruff}, they often fail to identify consistent and practically meaningful differences between algorithms~\cite{surveyzhao}. Furthermore, different studies rely on substantially different sets of algorithms, datasets, evaluation protocols, and hyperparameter settings~\cite{macrodata,metasurvey}, frequently leading to contradictory conclusions.

This raises important questions regarding the reproducibility and reliability of anomaly detection benchmarks. In this work, we therefore investigate the stability of algorithm rankings in anomaly detection and analyze the impact of common benchmarking choices on the resulting conclusions.
We will present our methods in the subsequent section and will then evaluate the impact of benchmark choices and benchmark components on rank instability. \footnote{To increase the reproducibility of our research, our code is freely available at \\ \href{https://github.com/psorus/reproduce}{github.com/psorus/reproduce}}




\section{Methods}

To study the reliability of anomaly detection benchmarks, we analyze how algorithm rankings change depending on benchmarking choices. For this purpose, we consider seven anomaly detection algorithms: KNN (k-nearest neighbors) ~\cite{knn}, LOF (Local Outlier Factor) ~\cite{lof}, IFOR (Isolation Forest)~\cite{ifor}, HBOS (Histogram-Based Outlier Score)~\cite{hbos}, PCA (Principal Component Analysis)~\cite{pca}, CBLOF (Cluster-Based Local Outlier Factor)~\cite{cblof}, and SEAN (Shallow Ensemble ANomaly detection)~\cite{sean}. We then evaluate how the relative ranking of these algorithms varies across different experimental settings.

We intentionally restrict our study to comparatively lightweight anomaly detection methods with low computational overhead. This choice is motivated by two considerations. First, our goal is not to identify the overall best-performing algorithm, but rather to investigate the sensitivity of benchmark conclusions to methodological choices. Second, limiting the computational complexity enables us to perform a large number of repeated benchmark evaluations across many datasets and settings.
As benchmark data, we use the 690 datasets provided by the OddBench benchmark suite~\cite{macrodata}. To reduce computational cost, especially for distance-based methods such as KNN and LOF, each dataset is limited to at most $5000$ randomly sampled training and test instances. Unless stated otherwise, we further select a random subset of $100$ datasets for each benchmark configuration.

To quantify the stability of algorithm rankings, we introduce our novel \emph{rank instability} metric. Let $\text{Rank}(\text{algo}, \text{setting})$ denote the rank achieved by an algorithm under a specific benchmark setting. We define the overall instability as the mean standard deviation of ranks across settings:

\[
\sigma_{\text{rank}} =
\mathrm{mean}_{\text{algo}}
\left(
\mathrm{std}_{\text{setting}}
\bigl(
\mathrm{Rank}(\text{algo}, \text{setting})
\bigr)
\right)
\]

A low $\sigma_{\text{rank}}$ indicates stable rankings across benchmark settings, whereas higher values indicate that rankings strongly depend on experimental design.

\vspace{1em}
Finally, benchmarking anomaly detection algorithms requires several experimental design choices. In this work, we focus on four major sources of variability:

\begin{itemize}
    \item[(1)] Dataset selection,
    \item[(2)] Evaluation metric (ROC-AUC~\cite{rocauc} or AUCPR~\cite{aucpr}),
    \item[(3)] Hyperparameter selection,
    \item[(4)] Random seed selection.
\end{itemize}

To measure the impact of each factor individually, we simulate $100$ random choices of the respective parameter while keeping all remaining components fixed.\footnote{For random seeds and hyperparameter configurations, we sample $10$ random values per algorithm and combine them randomly across algorithms to obtain $100$ distinct benchmark settings.} For each source of variability, we compute the resulting rank instability $\sigma_{\text{rank}}$. 



\section{Benchmark Choices and Benchmark Reliability}

\begin{figure}
    \centering
    \begin{tikzpicture}
        \pie[
            text=legend,
            radius=3,
            font=\small
        ]{
            44.9/KNN,
            12.3/LOF,
            17.4/IFOR,
            13.2/PCA,
            12.2/SEAN
        }
    \end{tikzpicture}
    \caption{Distribution of the best-performing algorithm across different benchmark settings. Most algorithms become the top-ranked method under at least some experimental configurations.}
    \label{fig:best}
\end{figure}
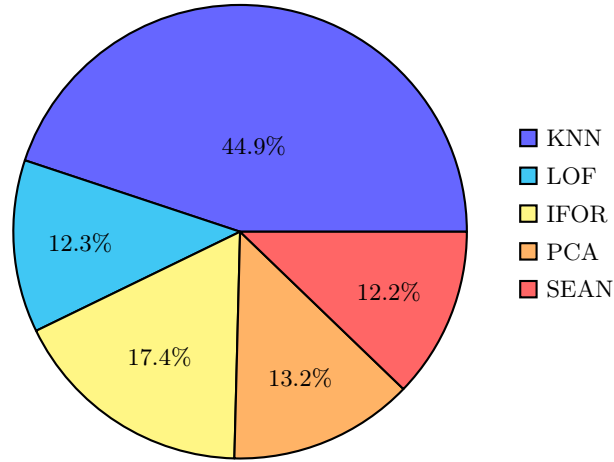

As we evaluate a large number of benchmark configurations, we observe that the ranking between algorithms changes substantially depending on the specific experimental design. In particular, varying the selected datasets, evaluation metrics, hyperparameter configurations, and random seeds can lead to entirely different conclusions regarding which algorithm performs best.
Figure~\ref{fig:best} illustrates the distribution of the best-performing algorithm across all evaluated benchmark settings. Remarkably, five of the seven studied algorithms achieve the top rank in more than $10\%$ of all settings. Only HBOS and CBLOF rarely appear as the best-performing methods. HBOS achieves the top rank in only one out of $1016$ evaluated settings, while CBLOF was never observed as the best-performing algorithm in our studies.

These observations have two important implications. First, they demonstrate that it is comparatively easy to make a reasonably strong anomaly detection algorithm appear state-of-the-art by selecting favorable benchmark settings. In our experiments, at most eight randomly sampled settings are sufficient to find a configuration in which a competitive algorithm achieves the best overall performance. Second, the results also indicate that this effect has practical limits: consistently underperforming algorithms cannot easily be made competitive solely through benchmark selection.
To better understand the origin of these ranking instabilities, we systematically investigate the impact of individual benchmarking choices on reproducibility and reliability in the following section.



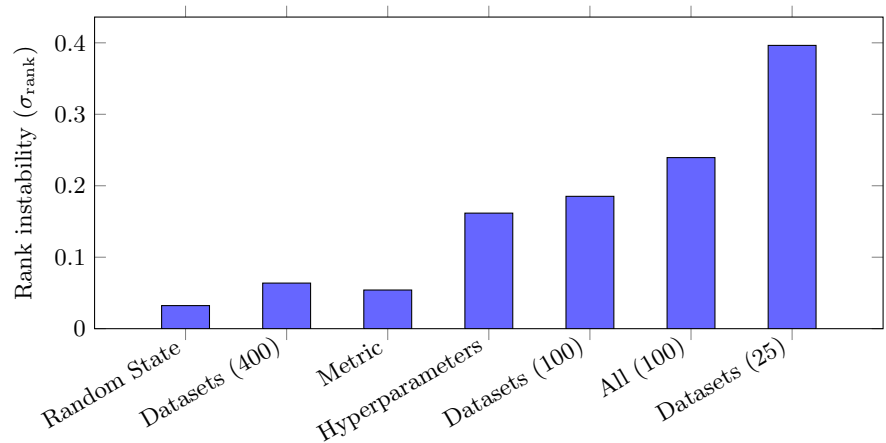
\begin{figure}[h!]
    \centering

    \begin{tikzpicture}
        \begin{axis}[
            ybar,
            bar width=18pt,
            width=12cm,
            height=5.7cm,
            ylabel={Rank instability ($\sigma_{\text{rank}}$)},
            symbolic x coords={
                Random State,
                Datasets (400),
                Metric,
                Hyperparameters,
                Datasets (100),
                All (100),
                Datasets (25)
            },
            xtick=data,
            x tick label style={rotate=30, anchor=east},
            nodes near coords align={vertical},
            every node near coord/.append style={font=\tiny},
            enlarge x limits=0.15,
            ymin=0
        ]
        
        \addplot[
            fill=blue!60,
        ] coordinates {
            (Random State,0.0323)
            (Datasets (400),0.0638)
            (Metric,0.0541)
            (Hyperparameters,0.1617)
            (Datasets (100),0.1852)
            (All (100),0.2394)
            (Datasets (25),0.3964)
        };
        
        \end{axis}
        \end{tikzpicture}
    
    \caption{Impact of different benchmarking choices on the rank instability $\sigma_{\text{rank}}$. Hyperparameter selection and dataset amount (25, 100, 400) contribute substantially more to ranking uncertainty than random seeds or evaluation metrics.}
    \label{fig:deltas}
\end{figure}

Figure~\ref{fig:deltas} shows the benchmarking results using our four sources of variability, namely dataset selection, evaluation metrics, hyperparameter selection, and random seed selection.
Our analysis demonstrates that benchmark rankings in anomaly detection are substantially affected by experimental design choices. At first glance, the observed instability of approximately $\sigma_{\text{rank}} \approx 0.34$ may appear moderate. However, under a Gaussian interpretation, this implies that only roughly $68\%$ of observed rankings lie within $\pm 0.34$ ranks of the expected ranking. Given the comparatively small ranking differences between algorithms, this uncertainty becomes highly significant in practice.
Indeed, across the seven studied algorithms, the average ranking varies by only $1.51$ positions overall, ranging from $3.37$ for KNN to $4.88$ for CBLOF. Consequently, an uncertainty of $0.34$ ranks represents a substantial fraction of the total observed performance gap between methods.

Importantly, not all benchmark components contribute equally to rank instability. The influence of random seeds is comparatively negligible, while the choice of evaluation metric has a smaller but still noticeable effect. In contrast, hyperparameter selection introduces substantial ranking uncertainty. Since our study focuses primarily on classical anomaly detection methods with comparatively limited hyperparameter spaces, this effect is likely still underestimated. Modern deep learning-based methods typically involve substantially more hyperparameters and are more sensitive to optimization choices, potentially leading to even larger reproducibility issues.

\begin{figure}[h]
    \centering

    \begin{tikzpicture}
        \begin{loglogaxis}[
            width=12cm,
            height=6cm,
            xlabel={Number of datasets},
            ylabel={Rank instability ($\sigma_{rank}$)},
            legend pos=north east,
            grid=both,
            major grid style={gray!20},
            minor grid style={gray!10},
        ]
        \addplot[
            only marks,
            mark=*,
            blue!60,
        ] coordinates {
            (1,1.9900998956279328)
            (2,1.360124144103289)
            (3,1.187025399980957)
            (4,1.0156595100018078)
            (5,0.8617760637011903)
            (6,0.8333577112467678)
            (7,0.7164471378016897)
            (8,0.7219122876112395)
            (10,0.6558524279348336)
            (11,0.5995430843585908)
            (13,0.569378749698477)
            (15,0.5118889351005468)
            (18,0.494166738411166)
            (21,0.44863657293701376)
            (25,0.4230006724821254)
            (29,0.4003106051933773)
            (34,0.38485764282153595)
            (39,0.36283772881427273)
            (46,0.3548749097509744)
            (54,0.3185433790656033)
            (63,0.304778658595312)
            (73,0.28062150532402624)
            (86,0.27003691174991346)
            (100,0.2566172890550786)
            (117,0.24129532615829424)
            (136,0.22868849631764426)
            (159,0.21619246086513044)
            (185,0.20800395293519114)
            (216,0.20612925597778622)
            (252,0.19253756010351072)
            (294,0.19003968841423277)
            (343,0.16644620342084368)
            (399,0.17619097691412255)
        };

        
        \addlegendentry{Observation}
        \addplot[
            domain=1:300,
            samples=200,
            thick,
            red!60,
        ] {1.63 * x^(-0.40)};
        
        \addlegendentry{$1.63 \cdot x^{-0.40}$}
        
        \addplot[
            domain=1:300,
            samples=2,
            dashed,
            red!60!blue!60
        ] {0.2};
        
        \addlegendentry{$\sigma_{rank} = 0.2$}
        
        \end{loglogaxis}
    \end{tikzpicture}
    
    \caption{Relationship between the number of datasets and rank instability $\sigma_{\text{rank}}$. Increasing the number of datasets improves ranking stability, but the effect saturates due to remaining sources of uncertainty.}
    \label{fig:odd}
\end{figure}
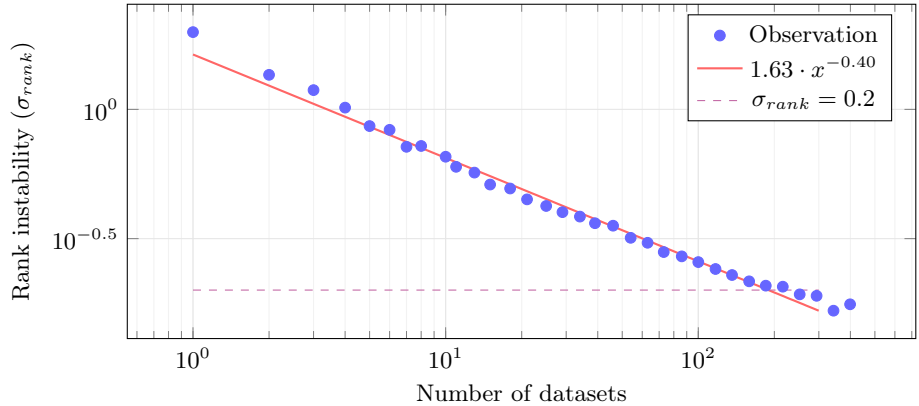

Finally, the number of datasets used in a benchmark has a major influence on ranking stability. We analyze this relationship further in Figure~\ref{fig:odd}. As expected, increasing the number of datasets reduces ranking uncertainty. However, this improvement eventually saturates, as the remaining sources of variability dominate the instability.
Overall, our experiments suggest that approximately $200$ datasets are required to reduce the total rank instability to a comparatively reliable level of around $\sigma_{\text{rank}} \approx 0.2$.

\section{Conclusion and Outlook}

In this work, we presented the first systematic study of how different benchmarking choices affect the reliability of anomaly detection benchmark studies. Our analysis demonstrates that benchmark conclusions in anomaly detection are highly sensitive to experimental design decisions and that algorithm rankings change substantially depending on the selected evaluation setup.

In particular, we showed that it is comparatively feasible to construct benchmark settings for most algorithms in which a newly proposed algorithm outperforms existing methods. At the same time, our experiments reveal that the dominant sources of ranking instability are the number of datasets used and the choice of hyperparameters. In contrast, factors such as random seed selection or the specific choice between common evaluation metrics contribute substantially less to the overall uncertainty.

Based on these findings, we recommend that future anomaly detection benchmark studies should include at least $200$ datasets whenever computationally feasible. Furthermore, benchmarks should explicitly evaluate algorithms under varying hyperparameter configurations instead of relying on a single manually selected setting and should include a value such as the herein proposed instability score. If computational resources are limited, increasing dataset diversity appears substantially more beneficial than evaluating additional random seeds or multiple closely related evaluation metrics.

More broadly, our results raise questions regarding the current emphasis on state-of-the-art performance in anomaly detection research. The observed ranking instability suggests that the notion of a universally ``best'' algorithm is often not robust and may depend strongly on benchmark construction choices. Consequently, we argue that future research should place greater emphasis on robustness, reproducibility, and understanding the conditions under which algorithms perform well, rather than focusing primarily on marginal improvements in benchmark rankings. 

\section*{Acknowledgment}
This research was in part funded by the Lamarr Institute for Machine Learning and Artificial Intelligence.





%
%
%
\bibliographystyle{splncs04}
\bibliography{refs,new}

\end{document}